\documentclass{edm_article}
\usepackage{multirow}
\usepackage{array}
\usepackage{url}
\begin{document}

\title{Analyzing Large Language Models for Classroom Discussion Assessment}

\numberofauthors{5}
\author{
%
%
\alignauthor
Nhat Tran\\
       \affaddr{University of Pittsburgh}\\
       \affaddr{Pittsburgh, PA, USA}\\
       \email{nlt26@pitt.edu}
\alignauthor
Benjamin Pierce\\
       \affaddr{University of Pittsburgh}\\
       \affaddr{Pittsburgh, PA, USA}\\
       \email{bep51@pitt.edu}   
\alignauthor
Diane Litman\\
       \affaddr{University of Pittsburgh}\\
       \affaddr{Pittsburgh, PA, USA}\\
       \email{dlitman@pitt.edu}
\and
\alignauthor
Richard Correnti\\
       \affaddr{University of Pittsburgh}\\
       \affaddr{Pittsburgh, PA, USA}\\
       \email{rcorrent@pitt.edu}
\alignauthor
Lindsay Clare Matsumura\\
       \affaddr{University of Pittsburgh}\\
       \affaddr{Pittsburgh, PA, USA}\\
       \email{lclare@pitt.edu}
}

 \maketitle

\begin{abstract}
Automatically assessing classroom discussion quality is becoming increasingly feasible with the help of new NLP advancements such as large language models (LLMs). In this work, we examine how the assessment performance of 2 LLMs interacts with 3 factors that may affect performance: task formulation, context length, and few-shot examples. We also explore the computational efficiency and predictive consistency of the 2 LLMs. Our results suggest that the 3 aforementioned factors do affect the performance of the tested LLMs and there is a relation between consistency and performance. We recommend a LLM-based assessment approach that has a good balance in terms of predictive performance, computational efficiency, and consistency.
\end{abstract}

\keywords{classroom discussion, large language models, scoring}

\section{Introduction}
Automatic assessment of classroom discussion quality has been a rising topic among educational researchers. Decades of research have shown that class discussion quality is central to learning \cite{talkmove_app, ian2015}. However, assessing classroom discussions in large numbers of classrooms has been expensive and infeasible to carry out at scale. Automated scoring of classroom discussion quality will aid researchers in generating large-scale data sets to identify mechanisms for how discussions influence student thinking and reasoning. In addition, automated scores could also be used in formative assessments (FA) to aid teachers in improving their discussion quality. 

The major advantage of modern large language models (LLMs) compared to pre-trained models such as BERT is that the former does not require training and only needs proper prompting to do the task. 
We attempt to test the capability of LLMs in automatically providing scores for different dimensions of classroom discussion quality, based on the {\it Instructional Quality Assessment (IQA)}, an established measure that has shown high levels of reliability and construct validity in prior learning research \cite{correnti_2}. 
Despite being new, LLMs have been used in classroom discussion assessments \cite{demszky-chatgpt, whitehill2024automated, chatgptstudent,finetunegpt}. However, prior work has largely used LLMs 
by designing a single prompt with fixed inputs and evaluating  zero-shot performance \cite{demszky-chatgpt, whitehill2024automated, chatgptstudent} or by finetuning which is costly and does not take advantage of the zero-shot or few-shot capability of LLMs \cite{finetunegpt}.
We instead analyze 3 factors that can potentially affect the predictive performance of the LLMs, as well as examine their impact on computational efficiency and consistency  in providing the same answer given the same input. 
Specifically, we test the capability of LLMs to score 4 IQA dimensions in various settings. 
First, different \textit{task formulation} in the prompt can be used depending on the way we formulate the task's goal \cite{taxonomy}. Second, unlike shorter inputs in other work on classroom discussion \cite{demszky-chatgpt, naz}, our transcripts are very long, which makes the \textit{context length} another factor worth testing as LLMs might not be able to process long-range context \cite{longrange}. 
Third, since LLMs are good few-shot learners \cite{fewshots}, we examine the utiliy of adding \textit{few-shot examples} to increase performance. 
Finally, we examine relationships between a LLM's {\it performance}, \textit{computational efficiency} and  \textit{consistency} in providing the scoring results.

Our contributions are two-fold. First, we show how 3 factors (task formulation, context length and few-shot examples) can influence LLM performance and computational efficiency in the IQA score prediction task.
Second, we examine the consistency between the LLMs' outputs 
and find correlations between performance and consistency in certain high-performance approaches. To support reproducibility, we also make our source code available at \url{https://github.com/nhattlm95/LLM_for_Classroom_Discussion}.

\section{Related work}
Researchers have measured classroom discussion at different grain sizes and with different foci. Human coding has often focused on either teaching moves or student moves, with some measures occurring at the utterance or turn level, while others focus on different dimensions of instructional quality using more holistic measures. Consequently, automated coding has followed similar directions \cite{alic-etal-2022-computationally, Jensen2020TowardAF,  demszky-etal-2021-measuring, demzsky2023-m, naz, suresh2021using, talkmove_app,   lugini-etal-2018-annotating, lugini-litman-2018-argument, lugini-litman-2020-contextual, paiheng2024a}.
{\it Our work focuses on automated holistic assessment of classroom discussion, both with and without also measuring fine-grained teacher and student moves.}

While most prior methods for automatically predicting talk moves and holistic scores \cite{talkmove_app,Jensen2020TowardAF,bert_model,jensen2021deep,naz, paiheng2024a}  have been based on modern NLP tools such as BERT \cite{bert}, recent work has started to explore the use of large language models (LLMs). 
For predicting accountable talk moves in classroom discussions, a finetuned LLM 
was shown to consistently outperform RoBERTa in terms of precision \cite{finetunegpt}. Since finetuning a LLM is costly and requires expertise, others have focused on zero-shot 
methods which do not require training. 
For example, the zero-shot capabilities of ChatGPT have been tested in scoring and providing insights on classroom instruction at both the transcript  \cite{demszky-chatgpt} and the sentence level \cite{whitehill2024automated, chatgptstudent}.  However, standard zero-shot approaches with fixed prompts were used and evaluated on pre-segmented excerpts of the transcripts, 
without further analyses of other factors that can potentially affect LMM performance. 
{\it Our work experiments with 3 such factors (i.e., zero versus few-shot examples, different prompting strategies, different input lengths)  that have
been shown to influence LLM performance  in other domains.}
First, different ways
of formulating a task in the prompt may yield different outcomes \cite{promptengineer, taxonomy, jiang-etal-2021-know}. Our study uses multiple prompting strategies reflecting different formulations of the holistic assessment task (e.g., end-to-end or via talk moves). 
Second,  LLMs can struggle in processing very long text input \cite{longrange}. Since our transcripts are often long,  we experiment with different ways of reducing the LLM input size.
Third, providing few-shot examples is known to be an effective way to increase LLM performance  \cite{taxonomy, liu-etal-2022-makes, fewshots, kojima2022large}. Since few-shot examples have not yet been utilized in previous classroom discussion LLM studies \cite{demszky-chatgpt, whitehill2024automated}, we propose a method for constructing such examples. 

In addition to testing the influence of task formulation, context length and few-shot examples on predictive performance, we also  \textit{evaluate the 3 factors' influence on computational efficiency (an important consideration for real-time formative assessment)}.
Finally, although aggregating multiple LLM's results for the same input (i.e., majority vote) has achieved higher performance in various NLP tasks \cite{self-consistency, programfc}, the consistency of the predicted results has not been examined in the context of classroom discussion. \textit{We explore result consistency at both the transcript and the score level and examine relationships with predictive performance}.
\section{Dataset}
Our corpus is created from videos (with institutional review approval)
of English Language Arts classes in a Texas district. 
18 teachers taught fourth grade,  13 taught fifth grade, and on average had 13 years of teaching experience. 
The 
student population 
was considered low income (61\%), with 
students identifying as: Latinx (73\%), Caucasian (15\%), African American (7\%), multiracial (4\%), and Asian or Pacific Islander (1\%).
The 
videos were manually scored holistically, on a scale from 1 to 4, using the \textit{IQA}  on 11 
dimensions \cite{lindsay_1}
for both teacher and student contributions. They were also scored using more fine-grained talk moves at the sentence level using the \textit{Analyzing Teaching Moves (ATM)}
discourse measure \cite{correnti_2}.  
The final corpus consists of \textbf{112} discussion transcripts that have already been converted to text-based codes (see Appendix \ref{app:data_stats} for the statistics of the scores). Thirty-two videos (29 percent) were double-scored indicating good to excellent reliability for holistic scores on the IQA (the Interclass Correlation Coefficients (ICC) range from .89-.98) and moderate to good reliability for fine-grained talk moves on the ATM (ICC range from .57 to .85). 
The university’s IRB approved all protocols such as for consent and data management (e.g., data collection, storage, and sharing policies). Privacy measures include anonymizing teacher names in the transcripts used for analysis.  Additionally, we only used open-source LLMs which do not expose our data to external sources.

The complete list of IQA dimensions can be found in Appendix \ref{app:other_iqa}. For this initial analysis, we focused on 4 of the 11 IQA dimensions. We chose these dimensions because of their relevance to dialogic teaching principles that emphasize collaborative knowledge-building and active participation in meaning-making processes. Two of the dimensions focus on teaching moves and 2 focus on student contributions. Furthermore, all 4 are calculated based on counting by their definitions.
We hypothesized that when combined these 4 dimensions would provide a theoretically grounded estimate of overall discussion quality. The four dimensions include:
(S1) \textit{Teacher links Student's contributions}, (S2) \textit{Teacher presses for information}, (S3) \textit{Student links other's contributions}, and (S4) \textit{Student supports claims with evidence}. We define S4 as max of (S4a) \textit{Student provides text-based evidence} and (S4b) \textit{Student provides explanation}. Descriptions of these dimensions can be found in Table \ref{tab:iqa}.
\begin{table*} [t!]
\centering
\caption{IQA dimensions and their definitions. For each IQA dimension (i.e., S1-S4), the italic line is \{IQA description\} and the remaining text is \{Scoring instruction\} used in the prompts in Section \ref{sec:methods}.}
\label{tab:iqa}
\begin{tabular}{p{2.4cm} | p{14.5cm}}

IQA Dimension & IQA Dimension's Description\\
\hline
 \textbf{S1. Teacher}
 
\textbf{links}
 
 \textbf{Student's}
 
 \textbf{contribution} &
 \textit{Did Teacher support Students in connecting ideas and positions to build coherence in the discussion about a text}?

4: 3+ times during the lesson, Teacher connects Students' contributions to each other and shows how ideas/ positions shared during the discussion relate to each other. 

3: Twice… 

2: Once… OR The Teacher links contributions to each other, but does not show how ideas/positions relate to each other (re-stating). 

1: The Teacher does not make any effort to link or revoice contributions. \\
\hline
 \textbf{S2. Teacher}
 
 \textbf{presses}
 
\textbf{Students}
 &
 \textit{Did Teacher press Students to support their contributions with evidence and/or reasoning}? 

4: 3+ times, Teacher asks Students academically relevant Questions, which may include asking Students to provide evidence for their contributions, pressing Students for accuracy, or to explain their reasoning.  

3: Twice… 

2: Once…OR There are superficial, trivial, or formulaic efforts to ask Students to provide evidence for their contributions or to explain their reasoning. 

1: There are no efforts to ask Students to provide evidence for their contributions or to ask Ss to explain their reasoning. \\
\hline
 \textbf{S3. Student}
 
 \textbf{links other's}
 
 \textbf{contributions} &
 
 \textit{Did Students’ contributions link to and build on each other during the discussion about a text}?

4:  3+ times during the lesson, Students connect their contributions to each other and show how ideas/positions shared during the discussion relate to each other. 

3: Twice… 

2:  Once… OR the Students link contributions to each other, but do not show how ideas/positions relate to each other (re-stating). 

1: The Students do not make any effort to link or revoice contributions. \\
\hline
\textbf{S4a. Student}

\textbf{provides}

\textbf{text-based}

\textbf{evidence} &

\textit{Did Students support their contributions with text-based evidence}? 

4: 3+ times, Students provide specific, accurate, and appropriate evidence for their claims in the form of references to the text. 

3: Twice… 

2: Once…OR There are superficial or trivial efforts to provide evidence. 

1: Students do not back up their claims. \\
\hline
\textbf{S4b. Student}

\textbf{provides}

\textbf{explanation} &

\textit{Did Students support their contributions with reasoning}? 

4: 3+ times, Students offer extended and clear explanation of their thinking. 

3: Twice… 

2: Once…OR There are superficial or trivial efforts to provide explanation. 

1: Students do not explain their thinking or reasoning. \\

\hline
\end{tabular}
\end{table*}

\begin{figure*}[t!]
\Description{Combinations}
\centering
\includegraphics[width=0.62\textwidth]{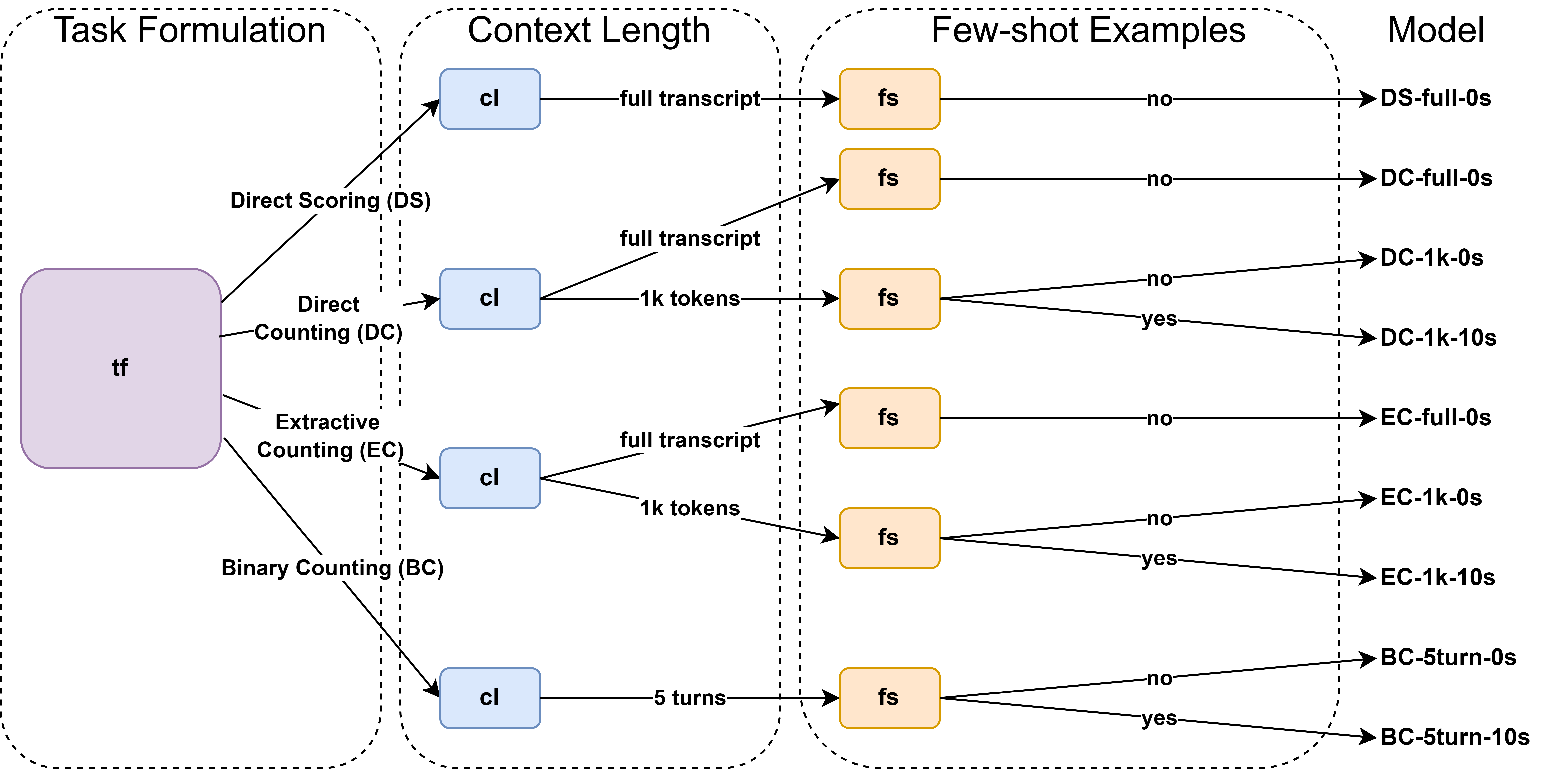}
\caption{The experimented LLM approaches and how they are constructed based on the 3 factors: Task Formulation (tf), Context Length (cl) and Few-shot Examples (fs).}
\label{fig:model_names}
\end{figure*}

\section{Methods}
Given a full classroom discussion transcript, our IQA score prediction task is to predict a score between 1 and 4 for each of the 4 targeted IQA dimensions. Because there are 3 factors that can affect the performance of LLMs, we use the same format to name the approaches. Specifically, each approach is named as \textbf{\textit{tf}-\textit{cl}-\textit{fs}} depending on the combination of the 3 factors: task formulation (\textit{tf}), context length (\textit{cl}) and few-shot examples (\textit{fs}). Figure \ref{fig:model_names} shows the final models and the combination that create them. Example prompts are in Apppendix \ref{app:example_prompts}.
In this section, we describe  3 factors and how we experimented with them in the task.
\label{sec:methods}
\begin{figure*}[t!]
\Description{Prompts}
\centering
\includegraphics[width=\textwidth]{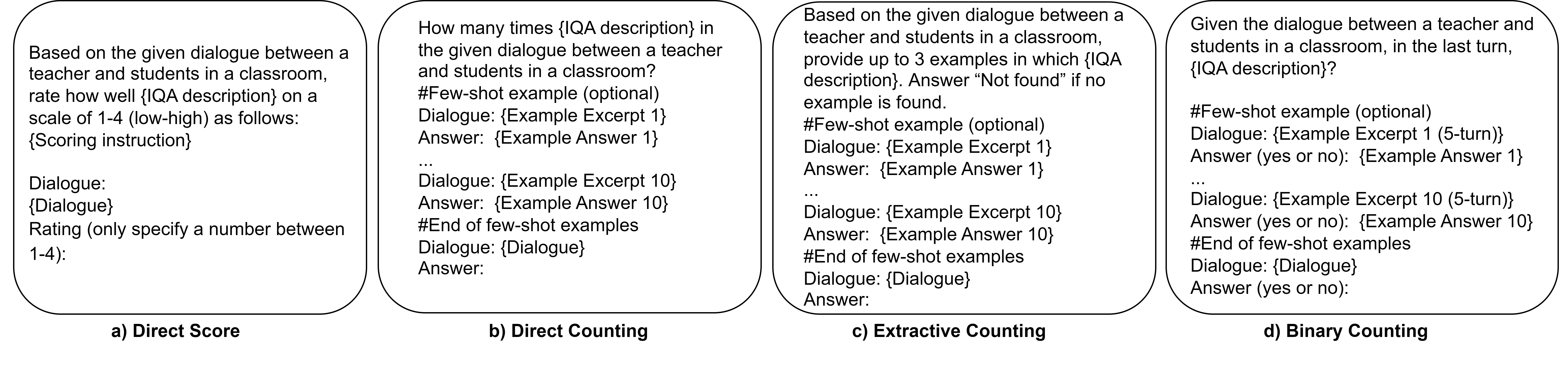}
\caption{Prompts used in this work. Lines starting with \# are comments and are not part of the prompts. \{IQA descriptions\} and \{Scoring instruction\} can be found in Table \ref{tab:iqa}.}
\label{fig:prompts}
\end{figure*}

\textbf{Task Formulation Factor}. 
LLMs receive instructions about the problem and how to achieve the desired results through prompts. Previous work has shown that different instructions can lead to different results for the same task \cite{promptengineer, taxonomy}. 
Additionally, although it is possible to prompt the LLM to do multiple tasks \cite{demszky-chatgpt}, our preliminary experiments show that the LLM sometimes fails to complete some or all of the tasks.
Therefore, we decided to use prompts that only require the LLM to do one task.
We experimented with the following 4 ways to formulate the task:

\textbf{Direct score (DS)}. We prompt the LLM to predict an IQA score for the transcript by giving it the description of each score for that dimension (1-4) (Figure \ref{fig:prompts}a). \{IQA description\} informs the LLM about the definition of the focused IQA score and \{Scoring instruction\} provides the criteria of each score from 1 to 4 for that IQA dimension. This is similar to end-to-end approaches that directly output the final score, either through transformer \cite{naz} or LLMs \cite{demszky-chatgpt, finetunegpt}.

\textbf{Direct counting (DC)}. For each IQA dimension, the description of each score from 1 to 4 is based on the count of relevant observations (i.e., a count of associated ATM codes at the turn level). Therefore, the \{Scoring instruction\} in DC can be formulated as a counting task.
We ask the LLM to count how many times a certain observation that represents an IQA dimension appears in the transcript by giving the IQA description (Figure \ref{fig:prompts}b). This can be treated as an alternative way to prompt the LLM with more direct and specific instructions (i.e., the LLM does not have to infer that the {Scoring instruction} is indeed a counting task).

\textbf{Extractive counting (EC)}. We prompt the LLM to extract turns from the transcript that satisfy certain observations that contribute to an IQA dimension (Figure \ref{fig:prompts}c). The final IQA score can be inferred by counting the number of turns found. This task formulation gives some explainability to the final score. Since a count higher or equal to 3 results in the maximum IQA score (4), we limit the number of extracted examples to 3 in the prompt.

\textbf{Binary counting (BC)}. We use the LLM as a binary classifier by prompting it to predict if an observation that represents an IQA dimension appears in one turn (yes/no) (Figure \ref{fig:prompts}d). Based on the performance in preliminary tests, we chose 4 previous turns for the dialogue history. 
Unlike the other 3 approaches which process the entire transcript in one go,
this approach uses LLM on the turn level. We then add the binary counts of each IQA dimension to get the final counts and infer the IQA scores.
This is similar to approaches identifying turn-level talk moves to predict holistic scores \cite{bert_model, naz}, except a LLM is the classifier instead of a transformer and there is no training/finetuning. This is also the most specific instruction as the output only has 2 labels (yes/no).

\textbf{Context Length Factor}.  
While previous work experimenting with LLMs on classroom discussion used short transcripts (e.g., several turns, 15-min passage) \cite{demszky-chatgpt, naz}, our transcripts are generally much longer (35 minutes on average). Specifically, 32 out of 112 transcripts have more than 4000 tokens, which exceed the token limit of many modern LLMs. Furthermore, although LLMs are claimed to be able to process long input text, their capabilities in dealing with long-range context are still questionable \cite{longrange}. Therefore, we test whether giving the LLM a shorter context such as an excerpt instead of the entire transcript (\{Dialogue\} in Figure \ref{fig:prompts}) leads to a change in performance.
For DC and EC, we split the transcripts into smaller excerpts of 1k tokens (best performance based on preliminary results) and aggregate the counts predicted by LLMs of each split to get the final counts of a transcript. 
We call these approaches \textbf{DC-1k} and \textbf{EC-1k}.

\begin{table*}[ht!]
\centering
\caption{Quadratic Weighted Kappa (QWK) from Mistral and Vicuna. The best numbers are bolded. Italic numbers mean they are equal or better than the BERT baseline. Inference time is the average of 3 runs.}\label{tab:qwk}
\begin{tabular}{c | l | c | c |c | c |c| c |c |c | c |c}

\multirow{3}{*}{ID} & \multirow{3}{*}{Method} & \multicolumn{5}{c|}{Mistral} & \multicolumn{5}{c}{Vicuna}\\
\cline{3-12}
 &  & \multicolumn{4}{c|}{IQA Dimension} & \multirow{2}{*}{Inference Time} & \multicolumn{4}{c|}{IQA Dimension} & \multirow{2}{*}{Inference Time}\\
\cline{3-6} \cline{8-11}
& & S1 & S2& S3 & S4 & & S1 & S2& S3 & S4 &\\

\hline
1&BERT \cite{bert_model} & 0.59 & 0.71 & 0.56 & 0.72 & 82s & 0.59 & 0.71 & 0.56 & 0.72 & 82s\\
\hline
2&DS-full-0s  & 0.42 & 0.50 & 0.43 & 0.49 & 10.7s & 0.42 & 0.48 & 0.45 & 0.48 & 12.3s\\
\hline
3&DC-full-0s  & 0.44 & 0.54 & 0.45 & 0.55 & 9.6s & 0.45 & 0.53 & 0.45 & 0.56 & 9.7s\\
4&DC-1k-0s & 0.46 & 0.57 & 0.47 & 0.61 & 10.7s & 0.47 & 0.57 & 0.49 & 0.60 & 12.5s\\
5&DC-1k-10s & 0.46 & 0.56 & 0.48 & 0.61& 12.3s & 0.46 & 0.58 & 0.49 & 0.63& 12.4s\\
\hline
6&EC-full-0s  & 0.43 & 0.58 & 0.45 & 0.60 & 17.1s & 0.41 & 0.54 & 0.42 & 0.60 & 18.2s\\
7&EC-1k-0s & 0.45 & 0.59 & 0.49 & 0.63& 24.7s & 0.45 & 0.57 & 0.47 & 0.64& 26.8s\\
8&EC-1k-10s & \textit{0.61} & \textit{0.71} & \textit{0.60} & \textit{0.74} & 27.3s & \textit{0.59} & 0.70 & \textit{0.56} & \textit{0.72} & 30.5s\\
\hline
9&BC-5turn-0s  & 0.49 & 0.62 & 0.50 & 0.65 & 223.4s & 0.49 & 0.63 & 0.50 & 0.63 & 234.1s \\
10&BC-5turn-10s & \textbf{\textit{0.63}} & \textbf{\textit{0.75}} & \textbf{\textit{0.64}} & \textbf{\textit{0.77}} & 232.5s & \textbf{\textit{0.62}} & \textbf{\textit{0.73}} & \textbf{\textit{0.60}} & \textbf{\textit{0.74}} & 237.9s\\
11 & \hspace{4mm} \textit{w/o} hard-negative & - & - & 0.55 & 0.69 & - & - & & \textit{0.56}& 0.69 & - \\
\end{tabular}
\end{table*}

\textbf{Few-Shot Examples Factor}.  
Providing examples is a simple yet effective way to improve a LLM's performance \cite{fewshots, taxonomy}. 
For approaches that have free spaces in the prompt, we try few-shot prompting by adding 10 more examples to the prompts. For BC, since each example is short (5 turns), we can freely provide any 10 5-turn excerpts with answers (yes/no) as few-shot examples for a selected IQA dimension. For DC-1K and EC-1K, we select 10 excerpts (700 tokens max) and infer their gold answers. The gold answer is the count of relevant ATM codes for DC-1K and a list of turns containing relevant ATM codes for EC-1K in the excerpt. We end up with 3 approaches using 10-shot examples: \textbf{DC-1k-10s}, \textbf{EC-1k-10s} and \textbf{BC-5turn-10s}.

For consistency, we have a fixed set of 10 examples for each approach.
To not expose test instances in these examples, we split the data into 2 segments A and B and for transcripts in one segment, we only draw examples from the other segment. In other words, for each IQA dimension of  DC-1k-10s, EC-1k-10s, and BC-5turn-10s, we create 2 10-example fixed sets (from segment A and B). When working on a transcript, only 1 of those 2 sets are used depending on the segment the transcript belongs to.
We also make sure that every possible label is covered in the 10 examples: 0-3 for DC, 0-3 extracted turns for EC, and yes/no for BC. 

To create those 10-example sets, instead of hand-picking the examples from the data, we use sampling. 
We use the word \textit{sample} from now on to describe the process of randomly selecting a text unit (several consecutive turns in the conversation) from the dataset until a certain condition is satisfied.
In BC-5turn-10s, for S1 and S2, we first sample 5 positive (yes) and then sample 5 negative (no) few-shot examples (5-turn each). For S3, S4a and S4b, since some negative examples are harder to distinguish from positive ones, we call them hard-negative examples. Specifically, they are turns containing the ATM code Weak Link (S3), Weak Text-based Evidence (S4a) and Weak Explanation (S4b). Previous work has shown that presenting hard-negative examples yields better prediction results \cite{learning-with-hard}.
Thus, we decide to sample 4 positive, 3 hard-negative and 3 easy-negative examples when predicting S3, S4a and S4b for BC-5turn-10s. 
For DC-1k-10s, we sample 2 text excerpts with the count of the IQA observation as \textit{k} (0 to 3), respectively, creating 8 examples. 
Similarly, for EC-1k-10s, we sample 2 dialogue excerpts in which \textit{k} (0 to 3) examples that satisfy the \{IQA description\} are extracted.
The last two examples of DC-1k-10s and EC-1k-10s do not have any restrictions.

\section{Experimental Setup}
\label{sec:es}
Commercial LLMs are costly and do not always guarantee data privacy, so we use open-source ones. To make a fair comparison with end-to-end scoring  (DS), we want a LLM that can fit long classroom discussion transcripts (as 32 out of 112 transcripts have more than 4000 tokens). Also, we want to test more than one LLM to make the findings more generalizable.
Among the open-source LLMs, Mistral \cite{mistral} and Vicuna \cite{vicuna} have a token limit of at least 8000, which is enough to cover any of our transcripts.
Specifically, we use Mistral-7B-Instruct-v0.1 \footnote{\url{https://huggingface.co/mistralai/Mistral-7B-Instruct-v0.1}} and Vicuna-7b-v1.5-16K \footnote{\url{https://huggingface.co/lmsys/vicuna-7b-v1.5-16k}} from huggingface with default parameters. 
We do not train or fine-tune the LLMs and use them as is.

To test the influence of the 3 aforementioned factors in the prediction, we report the average Quadratic Weighted Kappa (QWK) of the LLM approaches mentioned in Section \ref{sec:methods}. For BC-10s, we also report the performances on S3 and S4 without using hard-negative examples (i.e. 5 positives and 5 non-restricted negatives) to further test the effectiveness of having harder examples.
The baseline BERT-base model \cite{bert_model} was trained to predict the ATM codes by using either Hierarchical Classification (HC) for S1 or Sequence Labeling (SL) for S2, S3 and S4. The final IQA scores were inferred based on the counts of predicted ATM codes through a linear layer. Due to our small dataset, we use 5-fold cross-validation for this baseline, even though this makes the baseline not directly comparable to the 
zero-shot and few-shot approaches.
For each prompt, we run the LLM 3 times and aggregate the final predictions.
Since LLMs'outputs can be inconsistent, we use majority voting \footnote{We calculate the mean and round it to the closest integer if all 3 runs have different predictions} as previous work has shown that this is a simple yet effective technique \cite{self-consistency}.


To compare the computational efficiency, we record the average inference time (i.e., time to produce the set of 4 IQA scores for 1 transcript). 
We do not include the training time of BERT and the time spent on prompt engineering for LLMs. All experiments were done on a computer with a single RTX 3090 Nvidia GPU.

To measure the per-transcript consistency of LLMs, for each transcript, we record the number of times 2 out of 3 runs (2/3) and all 3 runs (3/3) have the same predictions per IQA dimension. The frequency that none of the 3 runs have the same predictions can be self-inferred. We also report the per-score consistency to see if the LLMs are more/less consistent in certain scores.

\section{Results and Discussion}

Table \ref{tab:qwk} shows the performances of the proposed approaches in Quadratic Weighted Kappa (QWK) along with their computational time for Mistral and Vicuna.

\textbf{Task formulation} is an important consideration as there were differences in performance on the IQA score assessment tasks. DS underperforms other approaches, including the baseline BERT model, with QWK scores of no more than 0.50 in all dimensions. This is consistent with a previous work which showed poor correlations between the scores predicted by a LLM and human raters on classroom transcripts \cite{demszky-chatgpt}. DC's variants (rows 3-5) outperform DS-full-0s, suggesting that the LLM cannot fully infer the relation between the counts of IQA observation and the final scores. EC-based approaches generally achieve higher QWK than DC's counterpart, except in some zero-shot instances (S1 of EC-full-0s and EC-1k-0s for Mistral; S1, S3 of EC-full-0s and EC-1k-0s for Vicuna). This implies that the LLM is generally better at extracting the IQA observations than counting them directly. 
The BC approaches obtain the highest performance, 
with BC-5turn-0s and BC-5turn-10s beating their counterparts (i.e., same few-shot settings) in all IQA dimensions, except for EC-1k-0s in S4 with Vicuna.

\textbf{Context length} 
also affects performance. 
With the same task formulation, reducing the context length to 1K always increases the QWK. 
BC-5turn-0s can be considered a zero-shot approach with a very short context length (5 turns) and it outperforms all other zero-shot approaches. These observations suggest that breaking a long transcript into smaller chunks of text is the recommended way when using LLMs for our task because it not only yields higher QWK but also enables usage of a wider variety of LLMs with lower token limits (e.g., LLama2 with a token limit of 4k).
\begin{figure*}[t!]
\Description{Per-transcript Consistency}
\centering
\includegraphics[width=\textwidth]{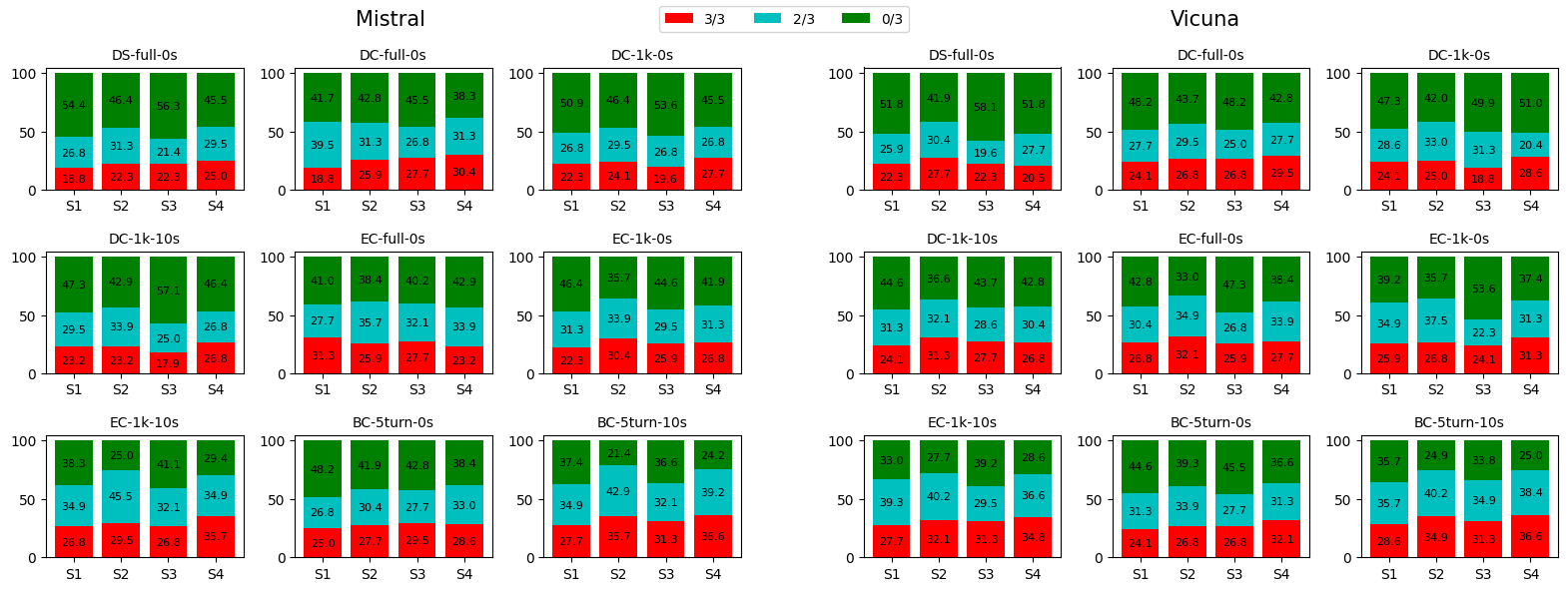}
\caption{Per-transcript statistics of the agreement from 3 runs for each approach. Each number is a percentage of the number of transcripts (out of 112). For each IQA dimension, 2/3 means exactly 2 out of 3 runs have the same predictions, 3/3 means all 3 runs have the same predictions, and 0/3 means no agreement between 3 runs.}
\label{fig:consistency_1}
\end{figure*}

\begin{figure*}[t!]
\Description{Per-Score Consistency}
\centering
\includegraphics[width=\textwidth]{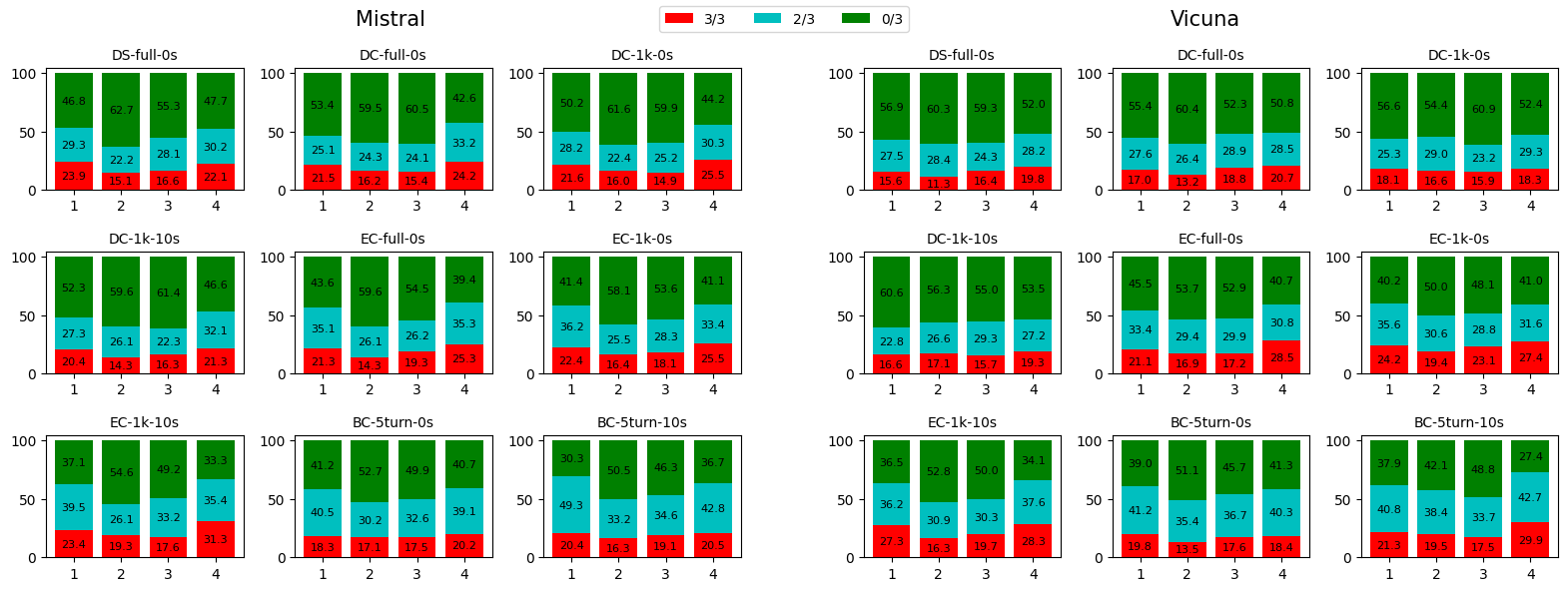}
\caption{Per-score statistics of the agreement from 3 runs for each approach. For each combination of a score \textit{s} (1-4) and an agreement rate (0/3, 2/3 or 3/3), each number is calculated as x/y where y is the number of times in any IQA dimension (S1-S4), the final prediction is \textit{s} and x is the number of times the agreement rate is satisfied among those y occurrences. For each score \textit{s}, 2/3 means 2 out of 3 runs have \textit{s} as predictions, 3/3 means all 3 runs have \textit{s} as predictions.}
\label{fig:consistency_2}
\end{figure*}
\textbf{Few-shot examples} do matter. The only two approaches that can outperform the baseline BERT model are both few-shot attempts (EC-1k-10s and BC-5turn-10s for both LLMs, except S2 in EC-1k-10s with Vicuna). 
The biggest gain in terms of performance is found when the Binary Counting approach is provided with 10 additional examples since BC-5turn-10s yields at least 0.10 points of QWK improvement over BC-5turn-0s, making it the best approach in all 4 IQA dimensions. 
While few-shot demonstration boosts the performances of Extractive Counting and Binary Counting, it does
not help Direct Counting since DC-1K-10s
performs similarly to DC-1K-0s, even worse in S2 with Mistral and S1 with Vicuna. We hypothesize that few-shot examples only help if they enhance the reasoning capability of LLM through those examples. For EC, the provided answers increase performance because the examples help the LLM better identify similar turns for scoring the IQA. For BC, the direct guidance from examples (yes/no) provides patterns (positive/negative) that the LLMs can absorb and generalize. In the case of DC, even with the correct counts given, the LLMs still need an intermediate reasoning step to identify the relevant IQA observations. In other words, the LLMs have to infer the characteristics of IQA observations from the counts - a task it struggles with. For DC, although the main task relies on counting, the bottleneck is likely from the capability of identifying related IQA observations, which the few-shot examples do not directly inject. The last two rows (10 and 11) also show BC-5turn-10s benefited from hard-negative examples, suggesting that having examples that are harder to distinguish from the focused labels when possible boosts classification performance of LLMs.

\textbf{Computational efficiency}.
The BERT approach runs slower than most of the LLM-based approaches except BC approaches because it processes on sentence level.
EC-based approaches run slower than DC-based approaches as the former require generating more tokens (generate a turn versus a single number).
BC approaches have superior performance in QWK compared to their counterparts but require excessive inference time. The best approach BC-5turn-10s needs around more than 8 times the amount of time to process a transcript on average compared to the second best approach EC-1k-10s. 
Although running slower, EC-based and BC-based approaches can be more useful if we want to go beyond summative to formative assessment for coaching or feedback as they present examples to justify the decision.
Therefore, if we want a balance between performance and inference time, EC-1k-10s is our recommended approach.

Figure \ref{fig:consistency_1} shows the transcript-level \textbf{consistency} across 3 runs for each approach. Although there are discrepancies among Mistral and Vicuna in different levels of agreement (2/3 and 3/3), most of the time, when majority voting is applied (i.e., at least 2 out of 3 agree on the final prediction), they are within 5\% of each other. 
The results also indicate that reaching total agreement (3/3) is hard for LLMs since the highest number is less than 37\%.
DS-full-0s is not only the worst approach performance-wise but also is very inconsistent as it has the lowest numbers overall (top 3 lowest agreement rates according to majority voting in all dimensions). On the other hand, the two approaches with the highest QWK, EC-1k-10s and BC-5turn-10s, obtain better consistency compared to the rest, especially in S2 and S4. Furthermore, similar to the QWK's result, S2 and S4 tend to have higher consistency than S1 and S3, suggesting that it is harder for LLMs to make consistent predictions on the latter dimensions. 
In general, these observations imply a relationship between performance and consistency of LLMs when the performance gaps are big, but when comparing approaches that are closer in performance,we see that an approach marginally better in QWK can have lower consistency (e.g., S3 of EC-1k-0s versus EC-full-0s). 

\begin{table}[t!]
\centering
\caption{Agreement rates for the score of 4 when we treat the exact counts as the prediction instead of rounding them down to 3 (score of 4) when the number of occurrences exceeds 3.}\label{tab:con_score_4}
\begin{tabular}{c | l | c c | c c }

\multirow{2}{*}{ID}& \multirow{2}{*}{Method} & \multicolumn{2}{c|}{Mistral} & \multicolumn{2}{c}{Vicuna}\\

&  & 2/3 & 3/3 & 2/3 & 3/3 \\
\hline
1& DC-full-0s & 25.4 & 20.2 & 26.3 & 16.7\\
2&DC-1k-10s & 28.4 & 18.3 & 28.1& 17.7\\
3&DC-1k-10s & 30.1 & 19.6 & 25.1 & 19.0 \\
4&EC-full-0s & 31.4 & 22.1 & 27.7& 21.9\\
5&EC-1k-0s & 29.3 & 20.2 & 28.5 & 23.6 \\
6&EC-1k-10s & 31.1 & 25.8 & 32.4 & 23.9\\
7&BC-5turn-0s & 34.6 & 17.4 & 33.7 & 16.4 \\
8&BC-5turn-10s & 34.9 & 17.5 & 36.4 & 25.4\\
\end{tabular}
\end{table}

Figure \ref{fig:consistency_2} reports the consistency across different scores. Overall, it is harder to reach a full agreement (3/3) for scores of 2 and 3 compared to 1 and 4 as all numbers in 3/3  for scores of 2 and 3 are lower than 20\% (except EC-1k-0s of Vicuna for score 3). BC-5turn-10s has the highest percentages in majority voting in general (sum of 2/3 and 3/3), and its consistency for scores of 2 and 3 is lower than for scores of 1 and 4. This suggests that the LLMs are more consistent when predicting the extreme scores (1 and 4). We hypothesize that because a score of 4 is correct whenever there are at least 3 occurrences of certain IQA observations, even if the LLM misses some occurrences, it can still predict 4 as the final answer if the total number of occurrences is large; or it can overcount but the final prediction is still 4 due to the rounding down. Table \ref{tab:con_score_4} supports this assumption because when we use the exact counts instead of limiting it to 3, we see a decrease in consistency for both 2/3 and 3/3 compared to Figure \ref{fig:consistency_2}. It implies that the LLMs are not very consistent for the score of 4 despite the high agreement rate from Figure \ref{fig:consistency_2}.
We leave further analyses to identify the problems of inconsistency for future work.

\section{Conclusion}
We experimented with 3 factors affecting the performance of 2 LLMs in the automated assessment of classroom discussion quality. Our results show that the 2 LLMs perform similarly and the task formulation is the most important factor that impacts the performance and inference time. A shorter context length generally yields higher results but requires more computational time. Furthermore, providing few-shot examples is a very effective technique to boost the performance of an LLM if it can utilize the cues from those examples. 
Further optimization on how to sample few-shot examples \cite{liu-etal-2022-makes, calibrate} is left for future work.
We believe in real-world applications, so finding a balance between the inference time and performance of LLM is crucial as it might not be worth sacrificing too much inference time for small performance gains. Finally, a brief count representing different levels of agreement across 3 runs shows that approaches that are noticeably better in prediction results are more likely to have higher consistency, but further analyses are still needed due to the overall low consistency. We would also like to examine how our findings generalize to other classroom discussion corpora and assessment schemes in future research.

\section{Limitations}
Due to our budget, we did not experiment with commercial LLMs such as GPT-4, which is more powerful and has a higher token limit. Additionally, although several other IQA dimensions can be tested using the same approach, we only worked on 4 of them.
Furthermore, human labor can provide better examples instead of choosing few-shot examples by random sampling from the data as we did. Despite its additional computational requirements, fine-tuning the LLMs, which has not been explored in this study, is a potential way to increase the performance further.
Since the experiments were conducted using a specific dataset (English Language Arts classes in a Texas district) and specific student demographics, a potential algorithmic bias might be present \cite{Baker2022}.

\section{Acknowledgments}
We thank the Learning Research and Development Center for the grant “Using ChatGPT to Analyze Classroom Discussions” and the Learning Engineering Tools Competition.
\bibliographystyle{abbrv}
\bibliography{EDM}  
%

\appendix

\section{Dataset Statistics}
Table \ref{tab:data} shows the statistics of the 4 focused IQA dimensions in our dataset.
\label{app:data_stats}
\begin{table*} [t!]
\centering
\caption{Data distribution and mean (\textbf{Avg}) of 4 focused \textit{IQA} rubrics for Teacher ($T$) and Student ($S$) with their relevant \textit{ATM} codes. An \textit{ IQA} rubric's distribution is represented as the counts of each score (1 to 4 from left to right) (n=112 discussions).}
\label{tab:data}
\begin{tabular}{l|c|c|c}
\multicolumn{3}{c|}{\textbf{IQA Rubric}} & \textbf{Relevant \textit{ATM} code}\\
\hline
\hfil \textbf{Short Description} & \textbf{Distribution} & \textbf{Avg Score} & \textbf{Code Label}  \\
\hline
S1: \textit{T} connects \textit{S}s & [69, 23, 9, 11] & 1.66 & Recap or Synthesize S Ideas\\
\hline
S2: \textit{T} presses $S$& [8, 13, 11, 80] & 3.46& Press \\
\hline
S3: $S$ builds on other's idea& [84, 7, 10, 11]& 1.54 & Strong Link \\
\hline
\multirow{2}{*}{S4: $S$ support their claims}& \multirow{2}{*} {[38, 17, 9, 48]} & \multirow{2}{*}{2.60}& Strong Text-based Evidence  \\
\cline{4-4}
& & & Strong Explanation  \\
\hline
\end{tabular}
\end{table*}

\section{Other IQA Dimensions}
\label{app:other_iqa}

We briefly list other IQA dimenions that were not studied in this work in Table \ref{tab:other_iqa}.

\begin{table*} [t!]
\centering
\caption{Other IQA dimensions that have not been studied in this work and their definitions.}
\label{tab:other_iqa}
\begin{tabular}{p{4.0cm} | p{13.0cm}}

IQA Dimension & IQA Dimension's Description\\
\hline
 Participation in Learning Community
 &
 {What percentage of Ss participated in the discussion about a text?}?\\




\hline
 Wait Time
 &
 {Did Teacher provide individual Students with adequate time in the class discussion to fully express their thoughts}? \\



\hline
 Rigor of Text &
 
 {How rigorous were the text(s) used as the basis for the discussion? Did they contain sufficient ‘grist’ to support an academically challenging discussion}? \\



\hline
Rigor of Class Discussion &

{Thinking about the text discussion as a whole and the questions Teacher asked Students, were Students supported to analyze and interpret a text (e.g., consider the underlying meaning or literary characteristics of a text, etc.)}? \\




\hline
Segmenting Text &

{Does Teacher stop during the reading of the text to ask questions or clarify ideas}?  \\





\hline
Guidance Toward Constructing the Gist &

{Does Teacher ask open-ended questions and facilitate discussion to guide Students to construct the gist of the text (i.e., a coherent representation of the text)}? \\




\hline
Developing Community &

{Does T help create a learning community within the classroom}? \\




\end{tabular}
\end{table*}

\section{Example Prompts}
\label{app:example_prompts}
\begin{figure}[t!]
\Description{DS Prompt}
\centering
\includegraphics[width=\columnwidth]{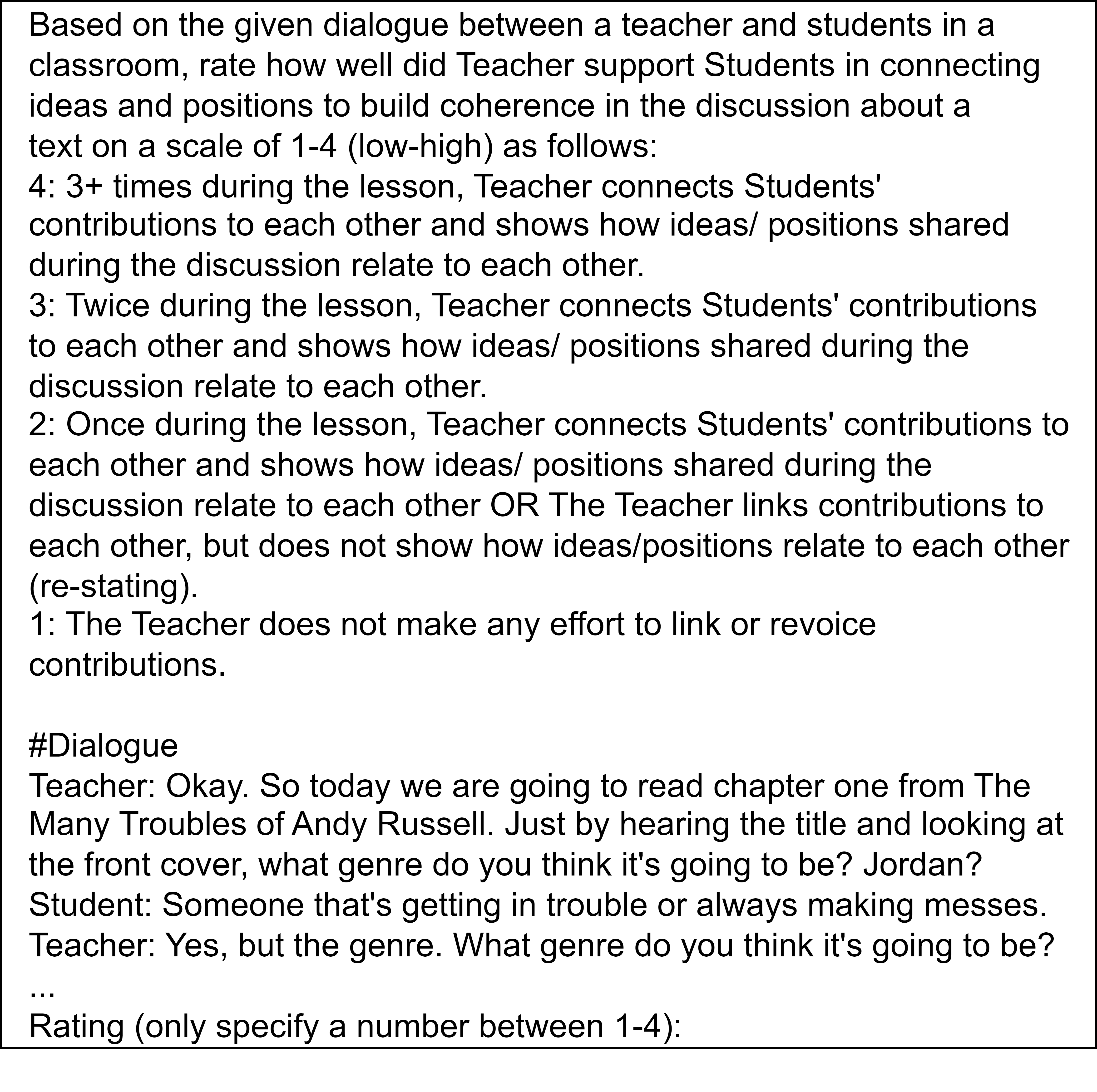}
\caption{An example prompt of Direct Score (DS) for (S1) Teacher links Student's contribution.}
\label{fig:prompt_ds}
\end{figure}

\begin{figure}[t!]
\Description{DC Prompt}
\centering
\includegraphics[width=\columnwidth]{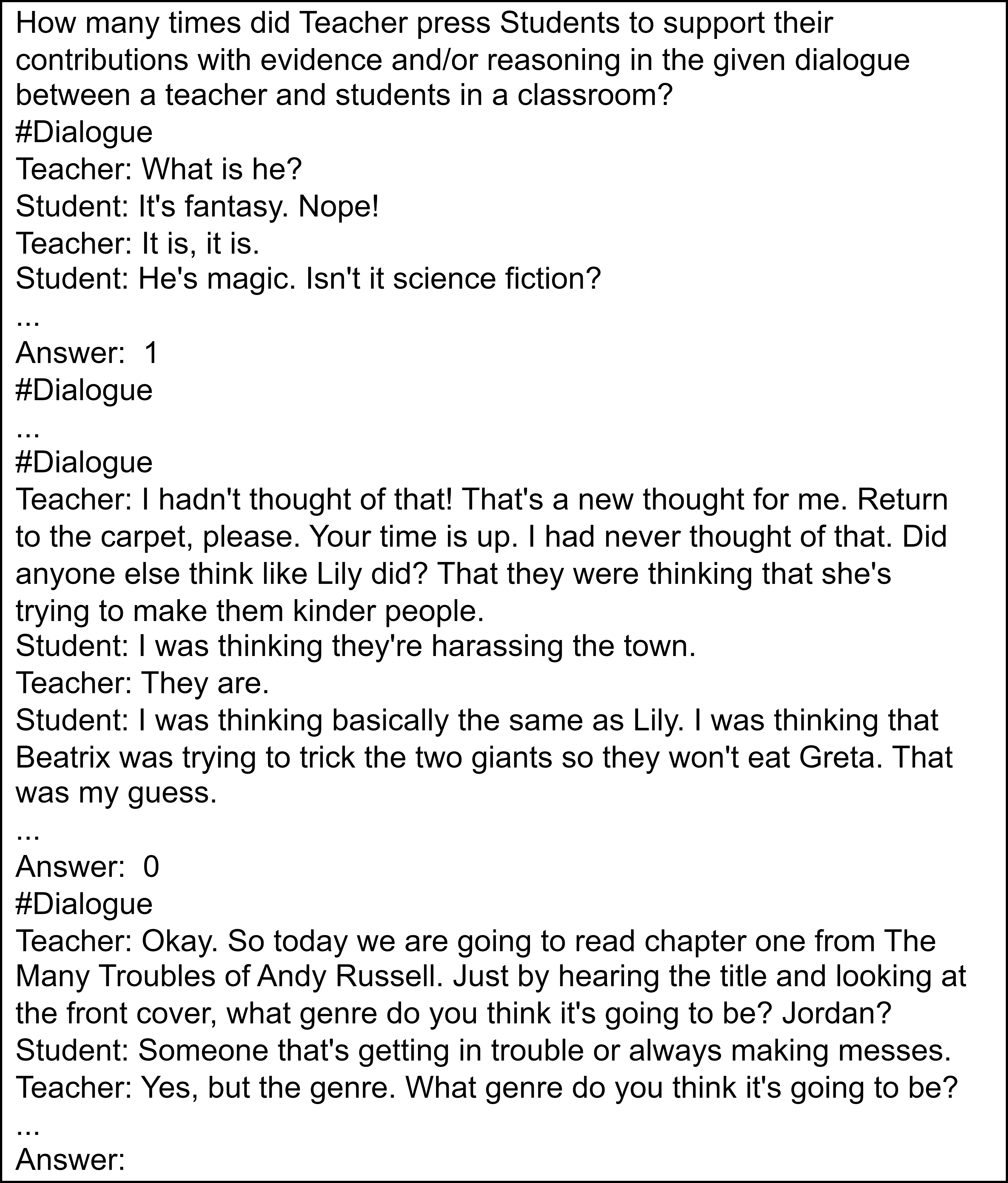}
\caption{An example prompt of Direct Counting (DC) for (S2) Teacher Presses Student.}
\label{fig:prompt_dc}
\end{figure}

\begin{figure}[t!]
\Description{EC Prompt}
\centering
\includegraphics[width=\columnwidth]{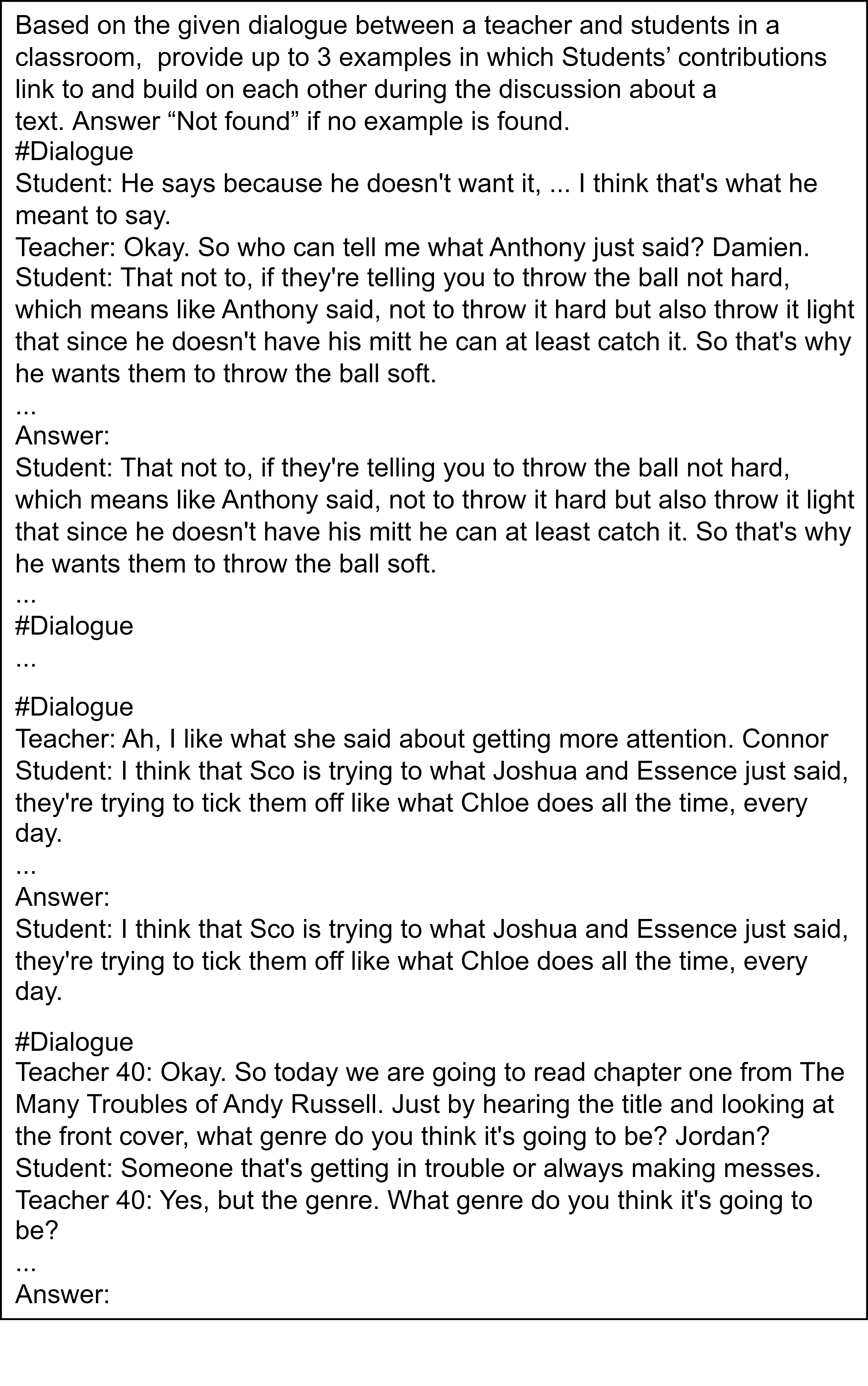}
\caption{An example prompt of Extractive Counting (EC) for (S3) Student links other's contribution}
\label{fig:prompt_ec}
\end{figure}

\begin{figure}[t!]
\Description{BC Prompt}
\centering
\includegraphics[width=\columnwidth]{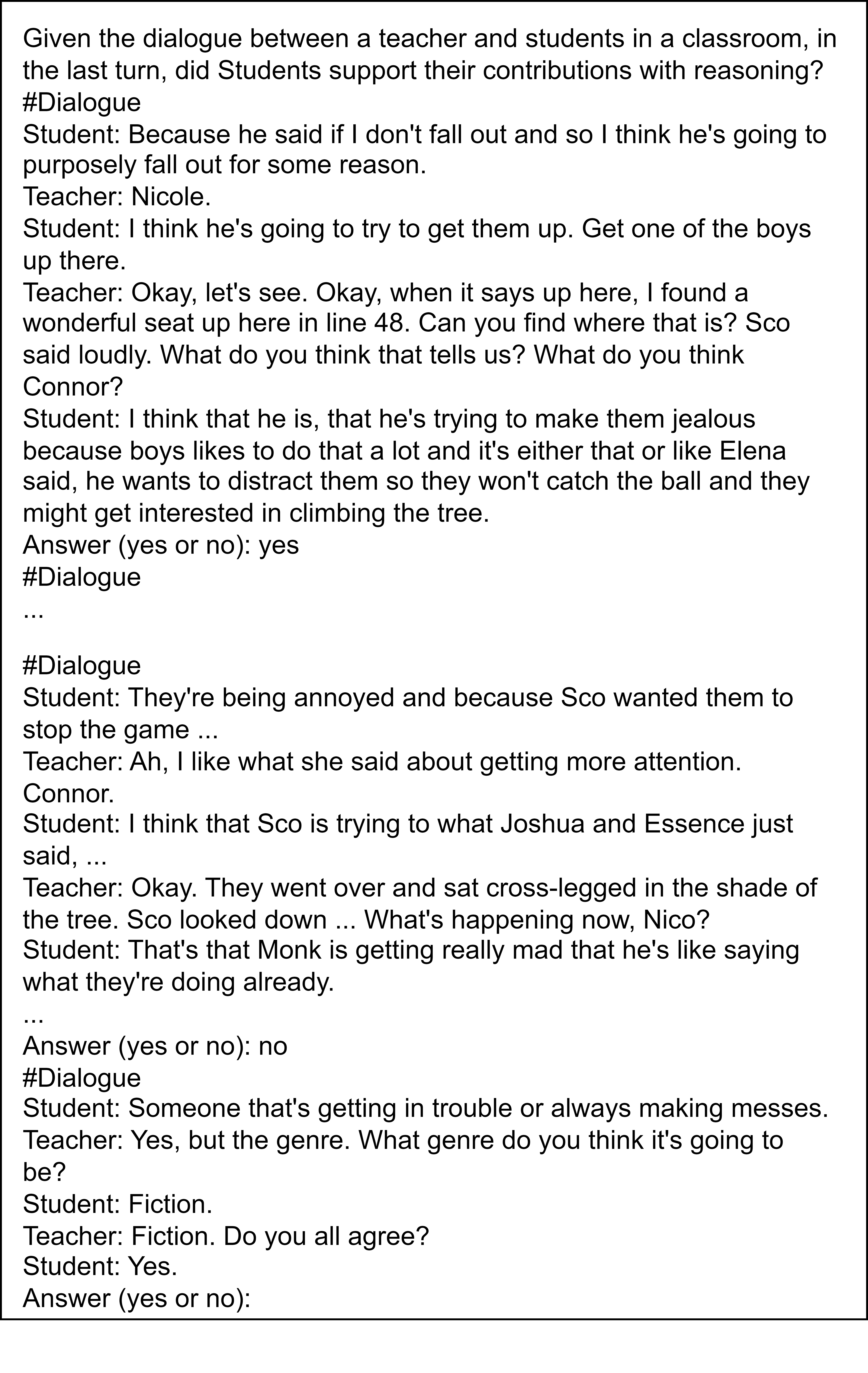}
\caption{An example prompt of Binary Counting (BC) for (S4a) Student provides explanation}
\label{fig:prompt_bc}
\end{figure}

Figures \ref{fig:prompt_ds}, \ref{fig:prompt_dc}, \ref{fig:prompt_ec} and \ref{fig:prompt_bc} show example prompts for Direct Score, Direct Counting, Extractive Counting and Binary Counting, respectively. The last lines of the prompts are incomplete to let the LLMs complete the text (i.e., provide the answer).

\end{document}